\documentclass[runningheads]{llncs}
\usepackage{graphicx}
\usepackage{epstopdf}
\usepackage{color}
\usepackage{subfigure}
\usepackage{booktabs}
\newcommand{\tabincell}[2]{\begin{tabular}{@{}#1@{}}#2\end{tabular}} 
\usepackage{url}

\usepackage{comment}

\begin{document}
\title{Travel Time Prediction using Tree-Based Ensembles}
%
%
\author{He Huang\inst{1}\and
Martin Pouls\inst{2}\orcidID{0000-0002-9258-719X} \and
Anne Meyer\inst{3}\orcidID{0000-0001-6380-1348} \and
Markus Pauly\inst{1}\orcidID{0000-0002-0976-7190}}
\authorrunning{Huang et al.}
%
\institute{TU Dortmund University, Department of Statistics, 44221 Dortmund, Germany \and
FZI Forschungszentrum Informatik, Information Process Engineering, 76131 Karlsruhe, Germany \and TU Dortmund University, Faculty of Mechanical Engineering, 44221 Dortmund, Germany
\email{he.huang@tu-dortmund.de}\\
}
\maketitle              

\begin{abstract}
In this paper, we consider the task of predicting travel times between two arbitrary points in an urban scenario. We view this problem from two temporal perspectives: long-term forecasting with a horizon of several days and short-term forecasting with a horizon of one hour. Both of these perspectives are relevant for planning tasks in the context of urban mobility and transportation services.
We utilize tree-based ensemble methods that we train and evaluate on a dataset of taxi trip records from New York City. Through extensive data analysis, we identify relevant temporal and spatial features. We also engineer additional features based on weather and routing data. The latter is obtained via a routing solver operating on the road network. The computational results show that the addition of this routing data can be beneficial to the model performance. Moreover, employing different models for short and long-term prediction is useful as short-term models are better suited to mirror current traffic conditions. In fact, we show that accurate short-term predictions may be obtained with only little training data.

\end{abstract}
%
%
%



\section{Introduction}


Predicting travel times of road trips is especially challenging in urban areas, as travel times considerably depend on the weekday, the time, and the current situation on the roads \cite{tomtom_2019,nyc_mobility_2016}. For both, the individual transport of people, and the transport of goods in urban areas, high quality predictions are necessary for planning reliable tours.
For operators of classic taxi services as well as providers of shared economy services such as Uber or Lyft a good prediction of arrival times is crucial, not only for creating efficient tours but also for having satisfied customers who do not have to wait longer than expected.
This is especially true if ride sharing services are offered, i.e., if different customers share one taxi in order to increase car utilization and decrease the price per person. 

Predictions for a time horizon of several days are necessary if guests book trips several days in advance and the taxi provider plans these tours ahead of time in order to guarantee a reliable service. Predictions of driving times for the next minutes or hour are necessary for dispatching, i.e., when assigning guests, who ask for an ad hoc service, to taxis on the short-term. Similar problems exist for other sectors such as same day deliveries of goods or food to consumers, but also deliveries to commercial customers such as restaurants, pharmacies, or shops which are often located in the city center. Hence, reliable travel time estimations in urban areas are necessary to (a) determine feasible plans, for example with respect to working time limits of drivers or opening hours of customers, and (b) to guarantee customer satisfaction.
A good overview of applications for planning vehicle routes in an urban context is given in \cite{Ulmer_2017}.

Historically, large traffic flow models (e.g., simulation, queuing theory) were built, calibrated with data from stationary devices such as induction loops or traffic cameras, to derive speed profiles for every section of the street network and to calculate the fastest route based on these profiles.
Today, GPS tracks -- referred to as floating car data -- are available from cars, and navigation systems. 
This considerably improves the availability of data for a larges share of road sections, and in some cases additional information such as speed data are available. 
Routing services such as Google directions\footnote{\url{https://developers.google.com/maps/documentation/directions/start} (13.05.2020)} or PTV Drive\&Arrive\footnote{\url{https://www.ptvgroup.com/en/solutions/products/ptv-driveandarrive/} (13.05.2020)} 
determine travel times based on these data for cars and trucks, respectively. 
Even if the travel time prediction of these services are of a high quality, the usage of a web API is often not possible: 
For dispatching decisions, the travel time for many relations is determined to assign a trip request to an adequate vehicle. 
If ride sharing services are offered, even more relations need to be considered.
Since dispatching decisions are extremely time sensitive, web service calls are prohibitive. Also the pay per call plans are usually expensive.
Finally, the predictions are provided for cars, pedestrians, or trucks but not for taxis, or small buses which are allowed to use special lanes in urban areas. In contrast to approaches based on complex traffic models estimating speed patterns on a link basis, we focus on simpler origin-destination based predictions: 
We only rely on information such as the location of the pickup and the deliveries and the corresponding time stamps.
These data are, usually, available even if not all cars are equipped with tracking systems.
If trajectories are available, selected nodes of the trajectory can be added as additional origin-destination data to the data base.
Due to the various advantages of tree-based learning methods, 
we investigate their suitability for predicting the travel times in the described setup.

The remainder of this paper is organized as follows:
In Section 2, we introduce the tree-based learning methods most appropriate to our problem at hand. 
Section 3 is dedicated to the computational study and evaluation of the methods for long-term prediction with a horizon of several days, and a short-term prediction for the next hours.
We conclude the paper with a short discussion. 
\section{Methodology: Tree-based Learning}

Travel time prediction can in principle be performed by any regression 
technique for metric outcomes. In particular, consider a theoretic model the form 
$$
y= f(x_1,\dots,x_p) +\epsilon =  f({\bf x}) +\epsilon,
$$
where $y$ denotes the travel time and $x_1,\dots,x_p$ are corresponding features, also called explanatory variables, containing, e.g., information on historic travel times, destination or even weather data which are stacked in the vector ${\bf x}=(x_1,\dots,x_p)$. Moreover, $\epsilon$ is an error term and $f$ an unknown regression function that needs to be estimated to describe the relationship between $y$ and $\bf x$. To this end, related applications on travel prediction have utilized approaches from time series analyses \cite{Guin_2006}
or Artificial Intelligence and Machine Learning. Here, support vector machines \cite{Wu_2004}, k-nearest neighbors \cite{Chang_2010} or neural networks and deep learning \cite{vanLint_2005,Duan_2016} have been proposed. We focus on methods that use trees as base learners. In particular, we study the performance of several tree-based methods such as CART and bagged or boosted ensembles. This includes {\it Random Forests} \cite{Breiman_2001}, {\it Extra (randomized) Trees} \cite{Geurts_2006} as well as the recently proposed {\it Xgboost} and {\it LightGBM} algorithms \cite{Chen_2016,Ke_2017}. These procedures are (i) known for being robust against feature co-linearity and high-dimensionality, (ii) usually more easy to train\footnote{at least compared to deep neural networks.} and (iii) also need a lower computational burden than more enhanced deep learning algorithms, see also \cite{Chen_2020,Yu_2018} for similar arguments. In fact, Random Forests or Stochastic Gradient Boosting models have already exhibit accurate predictions in the context of travel times \cite{Chen_2020,Yu_2018,Zhang_2015}; though for other data sets or research questions. For ease of presentation, we only summarize the basic ideas behind these methods and refer to the cited literature for the explicit definitions, see also the monograph \cite{Hastie_2009} for further details. As all methods are based upon trees, let us first recall the idea behind a single tree. 

\subsection{Classification and Regression Trees}
 Let $\mathcal{D}=\{(y_1,{\bf x}_1),\dots$ $,(y_n,{\bf x}_n)\}$ denote the observed data with ${\bf x}_i=(x_{i1},\dots,x_{ip})$ representing the feature vector of the $i$-th observation. We focus on regression trees from the {\it CART} class. 
The key idea is to greedily split the feature space into disjoint regions, say $R_1,\dots,R_m$, until a certain stop criterion is fulfilled. At the end, for each of the resulting disjoint regions, also called terminal nodes, a separate prediction of the target variable (travel time) is performed by 
\begin{equation}\label{eq:CART_pred}
 \widehat{c}_j = \frac{1}{N_j}\sum_{{\bf{x}}_i \in R_j} y_i,
\end{equation}
where $N_j=|\{\textbf{x}_i: \textbf{x}_i \in R_j\}|$. That is, the travel time of a new feature vector $\bf x\in R_j$ is predicted by taking the mean over all $y_i$ with feature vector ${\bf x}_i$ belonging to the  region that contains $\bf x$. 
To obtain the mentioned partition, binary splits are performed recursively, starting with the complete feature set (root node) and continuing with the resulting nodes etc. as follows: For each node, the observed feature value that minimizes the total variance of the two nodes after splitting is selected. A toy example is given in Figure~\ref{fig_CART} below, where we have chosen a tree depth of two. 
\begin{figure}[h!]
\begin{center}
\includegraphics[width=0.6\textwidth]{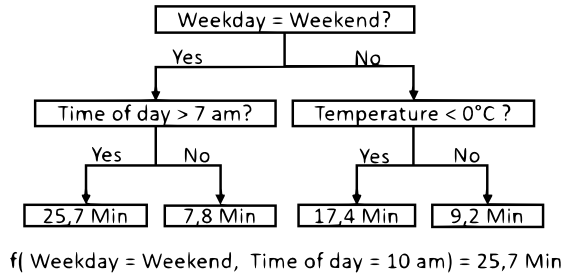}
\caption{A simple regression tree with tree depth equal to $2$.} \label{fig_CART}
\end{center}
\end{figure}
\vspace{-.5cm}
Here, the feature 'Weekday' with value 'Weekend' was chosen as first splitting point. Thus, two subsequent nodes partition the complete feature space into the observations belonging to Weekends (left node in the first row) and all other days (right node), respectively. Thereafter, the variable 'Time of day' with feature value '7 am' was chosen to split the data belonging to weekends. The two resulting sets are terminal nodes and the corresponding prediction values $\widehat{c}_1= 25,7\ \mathrm{Min}$ and $\widehat{c}_2= 7,8 \ \mathrm{Min}$ were calculated according to (\ref{eq:CART_pred}). Similarly, for weekdays the feature 'Temperature' with the freezing point as splitting value was chosen and $\widehat{c}_3 = 17,4 \ \mathrm{Min}$ and $\widehat{c}_4= 9,2\ \mathrm{Min}$ were calculated.\\

One of the most important questions with CART is the choice of the tree size. A strategy is to grow a large tree $B_0$ first, than reduce the size of the tree by using a pruning technique. Here, several pruning techniques exist and we have chosen to use the {\it Cost of Complexity Pruning (CCP)} introduced by Breiman \cite{Breiman_1984}. To describe this technique we define $B\subset B_0$ to be any sub-tree of $B_0$, which can be obtained by collapsing any number of its internal nodes. Let 
$
Q_j(B) =\sum_{({\bf{x}}_i, y_i) \in R_j}(y_i - \hat{c}_p)^2/{N_j}
$ 
be the mean squared error in region $R_p$. Then the CCP is defined as 
$
C_{\alpha}(B) = \sum_{j=1}^m N_j Q_j(B) + \alpha m,
$
where $\alpha$ is a tuning parameter for a trade-off between tree size and  goodness of fit. The sub-tree with a minimal $C_{\alpha}(B)$ will be selected, so that large $\alpha$ values lead to smaller tree sizes. 
This is an important point as a major advantage of trees is their interpretability as illustrated in the example from Figure~\ref{fig_CART}. However, if the tree size is too large, interpretation can become cumbersome. Moreover, trees come with the cost of a rather simple prediction model which can usually be enhanced in terms of accuracy by turning to ensemble techniques.



\subsection{Ensemble Techniques: Bagging and Boosting}
\begin{figure}[h]
\includegraphics[width=\textwidth]{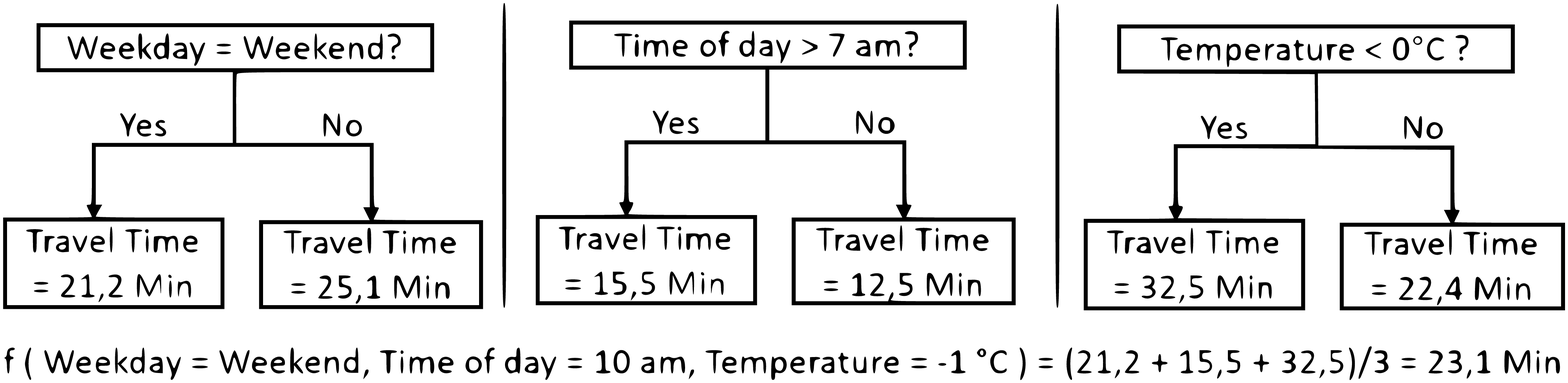}
\caption{A simple Random Forest with $B=3$ trees.} \label{fig_RF}
\end{figure}

Bagged ensembles as the Random Forest \cite{Breiman_2001} or Extra Trees \cite{Geurts_2006} usually improve the predictive accuracy of a single tree by following the wisdom of crowds principle: The basic idea is to randomly draw multiple subsamples, say $B$, from the (training) data and to grow a single tree for each of them and finally take the average tree prediction as random forest or Extra Tree prediction. To be a little bit more concrete, let $\mathcal{D}_b$ be a subsample (for the Random Forest this is usually of size $\lfloor.632n\rfloor$ and drawn with replacement from the training data $\mathcal{D}$) and ${\bf x}\mapsto \hat{f}_b({\bf x})$ the corresponding single tree predictor based upon $\mathcal{D}_b$, $b=1,\dots,B$. Then a bagged ensemble regression predictor is given by 
$
{\bf x}\mapsto  \sum_{b=1}^B \hat{f}_b({\bf x})/B = \widehat{f}({\bf x}).
$ 
The motivation behind this approach is that averaging reduces the variance. In fact, due to 
$$
Var(\widehat{f}) = \frac{1}{B^2} \left( \sum_{b=1}^B Var(\hat{f}_b) + \sum_{b_1\neq b_2} Cov(\hat{f}_{b_1},\hat{f}_{b_2}) \right)
$$
the trees should not be too correlated to further reduce the variance. That is why Random Forest as well as Extra Trees do not consider the same explanatory variables for each tree construction but draw random subsamples of feature variables. An example for the Random Forest is given in Figure~\ref{fig_RF}.
\vspace{-.5cm}
For simplicity, we only considered an ensemble of $B=3$ different stumps (trees). For each stump, we decided to randomly draw only one feature: The feature 'Weekday' for the first, the feature 'Time of day' for the second and 'Temperature' for the third tree. The numeric example below the trees explains the averaging principle. 
Random Forest and Extra Trees follow different approaches to generate (un-)corrolated trees. The key differences is that in Random forest the single trees are constructed as CARTs while the Extra Tree algorithm uses a different, even more random cut criterion to grow the trees. Roughly speaking, this should reduce the variance much faster (i.e. for smaller $B$) compared to Random Forests.

Different to these bagged ensembles, boosting is a more complex, iterative procedure. Therefore, we only explain some basic ideas and do not go too much into detail. One of its most common implementation is the gradient boosting algorithm \cite{Hastie_2009} which can be interpreted as a gradient descent implementation for the optimization of a loss function. Roughly speaking, it computes a sequence of trees, where each new tree is fitted on a modified version of the data (so-called pseudo-residuals) in order to reduce the value of a pre-specified loss function. Due to the iterative nature, the final model prediction is the sum from all tree predictions at each step. 
There exist several different boosting implementations which are all equipped with certain techniques for regularization and stopping rules. For our purposes we have chosen two boosting methods: Xgboost \cite{Chen_2016} as it has recently been crowned the current {\it Queen of machine learning} \cite{Morde_2019} due to its dominating performance in many applied Machine Learning and Artificial Intelligence competitions, as e.g., on Kaggle. In addition, we considered Microsoft's LighGBM which is advertised as being highly efficient and accurate \cite{Ke_2017}.

\section{Travel Time Prediction}

From common knowledge it may be apparent that travel time is affected by several variables such as the weekday, the time of the day, or the weather.
In this section we start describing the used data sets, and show selected results from  descriptive analyzes. 
For planning or dispatching transports, both, long-term predictions for travel times in several days, and short-term predictions for the next hour are necessary. 
Therefore, we consider different planning horizons and compare the resulting quality.
For the short-term prediction, we also evaluate how much data for training (the last hour or the last couple of hours) is needed to reach accurate short-term predictions with acceptable run times.


\subsection{Core data set and data enrichment}

As data set for our computational study, we use the ``TLC Trip Record Data'' containing trip information of the yellow and green taxis in New York provided by the NYC Taxi and Limousine Commission for the last 10 years \cite{tlc_2020}.
Since the information provided differs slightly between the years -- for example, the records provided later than 2016 do no longer contain location data -- we restrict our study to the period between January 2016 and June 2016.
To avoid data sparsity, we furthermore filtered our data set to trips which started and ended in Manhattan, which corresponds to the vast majority of trips.
In this period, the average number of trips per day is around 326 000, which results in an average number of trips per month of 9 949 000. 
Each trip 
is described by the date and the time, and the longitude and latitude value for its pickup and drop-off, respectively (in the following referred to as \emph{pickup$\_$datetime}, \emph{dropoff$\_$datetime}, \emph{pickup$\_$longitude}, \emph{pickup$\_$latitude}, \emph{dropoff$\_$ongitude}, \emph{dropoff$\_$latitude}).
Longitude and latitude values are given on 
five decimal places, which corresponds to an accuracy of about one meter.
Furthermore, the duration of the trips in seconds (\emph{trip$\_$duration}), the distance in miles (\emph{trip$\_$distance}), the number of passengers (\emph{passenger$\_$count}), and a code indicating the taxi provider (\emph{vendor$\_$id}) are provided.

\begin{figure}
    \centering 
    \includegraphics[width=.7\textwidth]{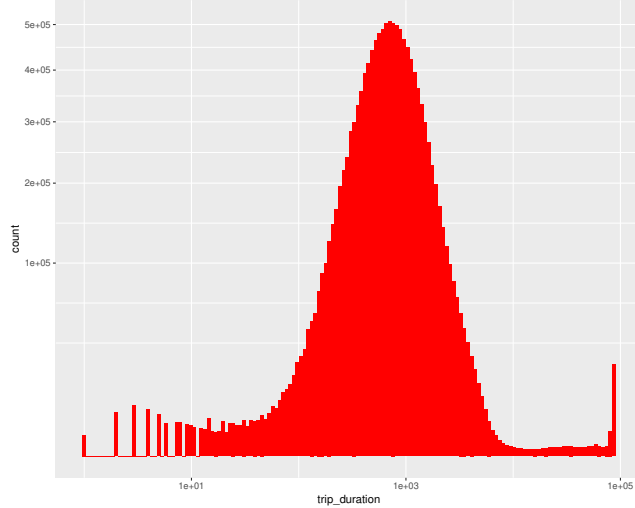}
    \caption{Distribution of the trip duration in seconds (x-axis on logarithmic scale).}
    \label{fig_DistrOfDuration}
\end{figure}

\begin{figure}
    \centering 
    \includegraphics[width=0.75\textwidth]{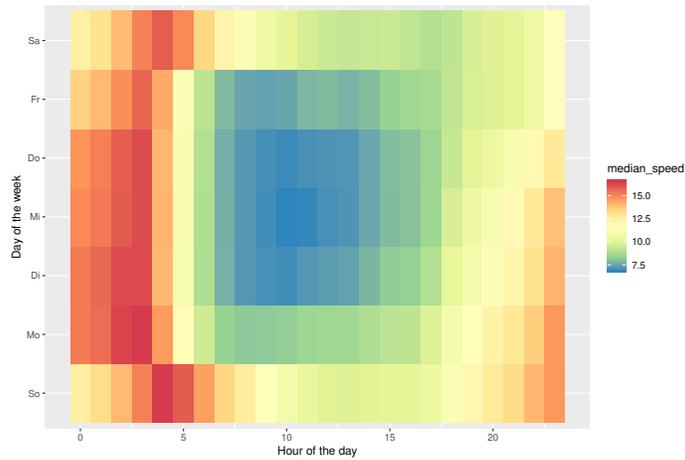}
    \caption{Interaction between weekday and hours on travel speed. The medians of the speed for each combination of weekday and hour are scaled with color. The combination of these two features created a “low speed region” in the middle of the day and week.}
    \label{fig_speedAndTime}
\end{figure}

\begin{figure}
    \centering
    \includegraphics[width=0.75\textwidth]{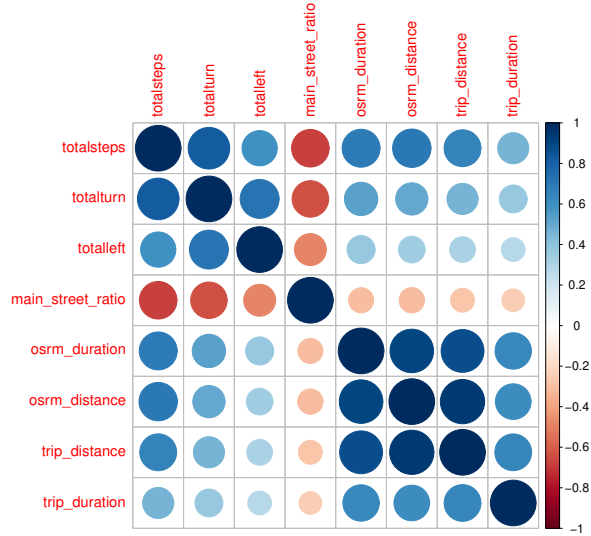}
    \caption{Pairwise correlations of the features.}
    \label{fig_correlation}
\end{figure}

Figure~\ref{fig_DistrOfDuration} shows the distribution of the trip duration, our target variable, for the area of Manhattan.
The distribution seems to follow a log-normal distribution with a peak of trips lasting around 16 minutes (around 1 000 seconds).
The trips with less than 10 seconds or with more than 100 000 seconds or more than 24 h
do not appear plausible. 
In this case we assume incorrect measurements.
\vspace{-.5cm}
The explanatory variables can be divided into three categories, namely temporal variables, spatial variables, and others. The temporal variables 
are the \emph{pickup$\_$datetime} and \emph{dropoff$\_$\-datetime}.
As expected, the day of the week, as well as the time of the day influence the travel time.
To understand this interaction we plotted the travel speed for each combination of "weekdays" and "hours of the day" in Fig.~\ref{fig_speedAndTime} with a heat-map. 
It shows that the combination created a “low speed region” in the middle of the day and week.
Such speed profiles over the course of the days are typical for urban areas, and they correlate negatively with the number of taxi trips. 
The speed goes down with an increase of road users during the day.
During the night the speed, usually, corresponds to the so called free flow speed and is close to the legal speed limit for each road section.
We also investigated the influence of weather on the speed using data provided by the National Weather Service\footnote{The data can be downloaded from \url{https://www.weather.gov}}.
The impact of heavy snowfalls in January on the travel speed was significant.
Other weather phenomena, such as rainfall, do not have a similar impact on the travel speed.
However, the reason can also be the shortcoming of our weather data set: 
It contains only average values per day, even though rainfall changes considerably over the course of the day.  
The spatial variables in our data set are the longitude and latitude values for the pick-up and the drop-off, and the distance of the trip. As shown in Fig.~\ref{fig_correlation}, the \emph{trip$\_$distance} has a strong positive correlation with our target variable \emph{trip$\_$duration}.
The actual trip distance can only be measured ex-post, since the traffic situation influences the concrete route, and often taxi drivers use their geographical knowledge and use shortcuts.
However, we can calculate the fastest route between the pick-up and the drop-off location with a routing algorithm assuming free flow speed on all road sections and use the resulting distance as an estimator for the trip distance which is available ex-ante.
For our study, we applied the Open Source Routing Machine (OSRM)\footnote{\url{http://project-osrm.org/}} on open street map data for New York.
Besides the distance, referred to as \emph{osrm\_distance}, the response generated by the OSRM engine also contains the following information for the fastest (not necessarily shortest) route: 
the duration (\emph{osrm\_duration}), the total number of left-, right-, and u-turns (\emph{total\_turns}), the total number of left-turns (\emph{total\-\_left}), the total number of navigation instructions (\emph{total\_steps}), the name of the street with the longest duration (\emph{main\_street}), and the fraction of travel time on the main street (\emph{main\_street\_ratio}).

Fig. \ref{fig_correlation} shows that the \emph{osrm$\_$distance} and \emph{osrm$\_$duration} of the fastest route are highly correlated with the actual duration and distance. 
The \emph{main$\_$street$\_$ratio} is weekly (negative) related to the actual trip duration. 
The \emph{total$\_$steps}, \emph{total$\_$\-left}, \emph{total$\_$turns} and \emph{main$\_$street$\_$ratio} from OSRM are highly correlated with each other. 
Thus, it's not necessary to use all these features for prediction.\\

\subsection{Long-Term Forecasting}

\subsubsection{Cleaning.}

\begin{figure}
\includegraphics[width=0.9\textwidth]{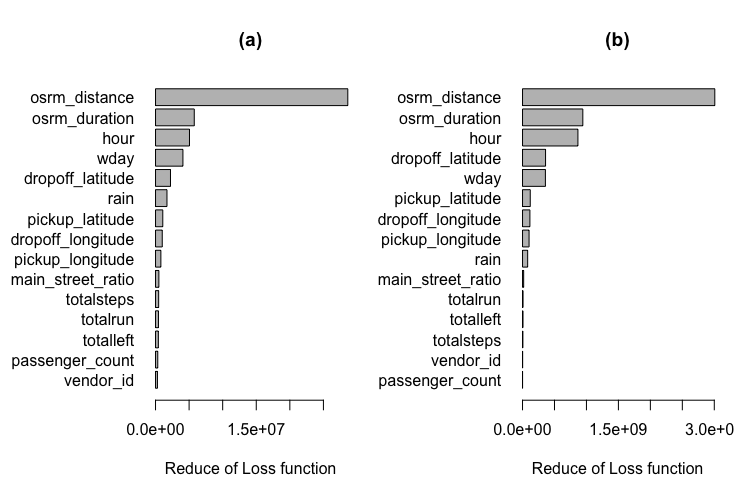}
\caption{Feature importance of (a) XgBoost and (b) lightGBM.}
\label{fig_featureImportance}
\end{figure}

Each model was trained on data from 1st June 2016 to 20th June 2016 (training data) and 
its prediction ability was evaluated in terms of RMSE on the data form 21st June 2016 to 30th June 2016 (test data). The training and test data contain 6,355,770 and 3,105,839 trips respectively, with $16$ features for each trip. In Machine Learning, it is usually recommended to clean up the data before training. To this end, we first removed all trips with untrustworthy feature values, e.g., 10,000 trips with less than 10 seconds or more than three hours duration within Manhattan as well as 2,000 trips with an average speed of more than 60 miles per hour (noting that Manhattan's maximum speed limit is 50 miles per hour). In addition, we did a descriptive analysis of all feature variables to identify features without significant impact. Beyond summary statistics and graphical illustrations (not shown) we thereby also calculated the feature importance of all untuned models on the trip duration. Due to similarity we only show the resulting plots for the two boosting approaches in Figure~\ref{fig_featureImportance}. 
Here, \emph{vendor$\_$id} and \emph{passenger$\_$count} had the least impact. As the exploratory analyses (bar- and boxplots) agree with this assessment, we excluded both from our further analyses. Moreover, \emph{total$\_$steps}, \emph{total$\_$left}, \emph{total$\_$turns} and \emph{main$\_$street$\_$ratio} from OSRM express the next lowes feature importance. As Figure~\ref{fig_correlation} already revealed that these four features were highly correlated, we only keep the \emph{main$\_$street$\_$ratio} and removed the other three. 

Based on the remaining training data on these $p=11$ features ${\bf x}$ and the target variable $y$, we performed an independent hyperparameter tuning for each of the five different modeling approaches. These were done on Python \cite{team_2019}. In particular, 
for CART and Extra Trees we used the library scikit-learn \cite{Pedregosa_2011}, for Random Forest and XgBoost the XgBoost-library from \cite{Chen_2016}\footnote{The source code of this library is available at \url{https://github.com/dmlc/xgboost}} was utilized and for lightGBM we turned to the Microsoft library\footnote{The source code is available at \url{https://github.com/microsoft/LightGBM}}. As computational complexity can become an issue with boosting algorithms, we additionally compared the performance of the two boosting algorithms on both, CPUs and GPUs. For the latter, we turned to the plug-in GPU implementation of \cite{Mitchell_2017}. As this library only supports the faster histogram-based implementation of the XgBoost algorithm, 
we also restricted the CPU based XgBoost-implementation to this. This leaves us with seven different models to train. 
\subsubsection{Parameter Tuning.} Starting with the single decision tree CART we tuned the size of the tree by \emph{max$\_$depth} and \emph{min$\_$child$\_$weight}. 
To this end, we run a grid search for \emph{max$\_$depth} $\in \{3, 8, 13, 18, 23, 38, 33\}$ and \emph{min$\_$child$\_$weight} $\in \{10, 20, ..., 130\}$ resulting in the choices \emph{max$\_$depth} $=23$ and \emph{min$\_$child$\_$weight}$=100$. 
For Random Forest, the main parameters to be tuned are the subsampling rate on the training data (\emph{subsample}), the subsampling rate on the features when building each tree (\emph{colsample$\_$bytree}) and the total number of trees (\emph{n$\_$trees}). 
Apart from these bagging parameters, we also tuned the value \emph{max$\_$depth} for each single tree. We created a grid for these four parameters with \emph{subsample} and \emph{colsample$\_$bytree} in $\{0.6, 0.7, 0.8, 0.9, 1.0\}$, \emph{n$\_$trees} $\in\{20, 40, 60, 80, 100\}$ and  \emph{max$\_$depth} $\in\{8, 13, 18, 23, 28\}$. 
The resulting best parameters with a minimal training RMSE of $254.21$ were \emph{max$\_$depth}$=28$, \emph{colsample$\_$bytree}$=0.8$, \emph{subsample}$=0.9$ and \emph{n$\_$trees}$=80$. A similar grid search for the Extra Trees parameters (for which \emph{subsample} is redundant) with \emph{colsample$\_$bytree} $\in\{0.6, 0.7, 0.8, 0.9, 1.0\}$, \emph{n$\_$trees} $\in\{20, 40, 60, 80, 100\}$ and \emph{max$\_$depth} $\in\{ 3, 8, 13, 18, 23, 28, 33\}$ led to a minimal RMSE of $261.3$ for the choices \emph{n$\_$trees}$=60$, \emph{max$\_$depth}$=28$ and  \emph{colsample$\_$bytree}$=0.6$. Compared these bagging method, XgBoost and lightGBM 
needed $8$ parameters to tune. As an $8$-dimensional grid was computationally too exhaustive with more than $72$ Mio values ($6$ Mio for the target variable and each of the $11$ features), 
we performed a step by step approach in which step we only tuned some of them with grid search.

\subsubsection{Model Comparison.} The performance of the resulting 'best' tree-based models were compared with respect to their predictive accuracy (measured in RMSE) on the test set and their computational effort needed for training. For a baseline comparison, a naive prediction was computed which uses 
the average trip duration for each combination of pickup and drop-off zipcode under the same same hour of the same week day. 
The results are summarized in Table~\ref{tab_comptreebase}. Regarding computational time we considered two situations: The training time to obtain the corresponding 'best' model (second last column) and for illustration also the training time under the same tree and ensemble sizes (last column) 
\emph{max$\_$depth}$=16$ and \emph{n$\_$trees}$=60$. All computations were performed on an Intel Core i7-8700 3.20GHz $\times$ 12 (CPUs) with GeForce GTX 1080/PCIe/SSE2 (GPUs) and 48 GB RAM. 

\begin{table}[htb]
\centering
\begin{tabular}{p{0.23\textwidth} p{0.13\textwidth} p{0.3\textwidth} p{0.3\textwidth}}
\toprule
Model             & RMSE     &  \tabincell{c}{Training Time [s] \\ of best Model} &  \tabincell{c}{Training Time [s] \\ under same parameters} \\ \midrule
Naive             & 368.97   &  -                               & -    \\
 CART             & 271.23   &  72                              &  -    \\ 
Random Forest     & 254.21   &  442 (with 80 trees)             &  442  \\
Extra Trees         & 261.3   &  825.6                          &  493 \\
XgBoost(GPU) & 253.74   &  5916 (with 1685 trees)          &  72\\                     
XgBoost(CPU) & 253.37   &  7240 (with 1662 trees)          &  196 \\ 
lightGBM(GPU)     & 256.59   &  139 (with 597 trees)             &  9 \\
lightGBM(CPU)     & 256.46   &  159 (with 524 trees)             &  11\\ 
\bottomrule
 \end{tabular}
 \caption{Comparison of the Tree-based Models. Some of the models, whose parameter haven't been tuned, have no data for the best models.}
\label{tab_comptreebase}
\end{table}\vspace{-.5cm}
It is apparent that the performance of all tree-based ensemble models were much better than the naive method and also better than the single CART tree. Among them, XgBoost on CPU was the model with the best RMSE (253.37) on test data directly followed by its GPU implementation (253.74). Moreover, Random Forest (254,21) and both lightGBM implementations ($\approx 256$) only showed slightly worse accuracies while needing much less time for training. In fact, turning from XgBoost on its faster GPU implementation to the well-established Random Forest reduced the computational burden to train the best model by the factor $13.5$. Turning to lightGBM (GPU implementation) even resulted in the factor $42.9$.
Finally, Extra Trees exhibits the worst RMSE among all ensembles while needing more time to train than lightGBM and the Random Forest.



\subsection{Short-Term Forecasting}

\begin{figure}
\includegraphics[width=\textwidth]{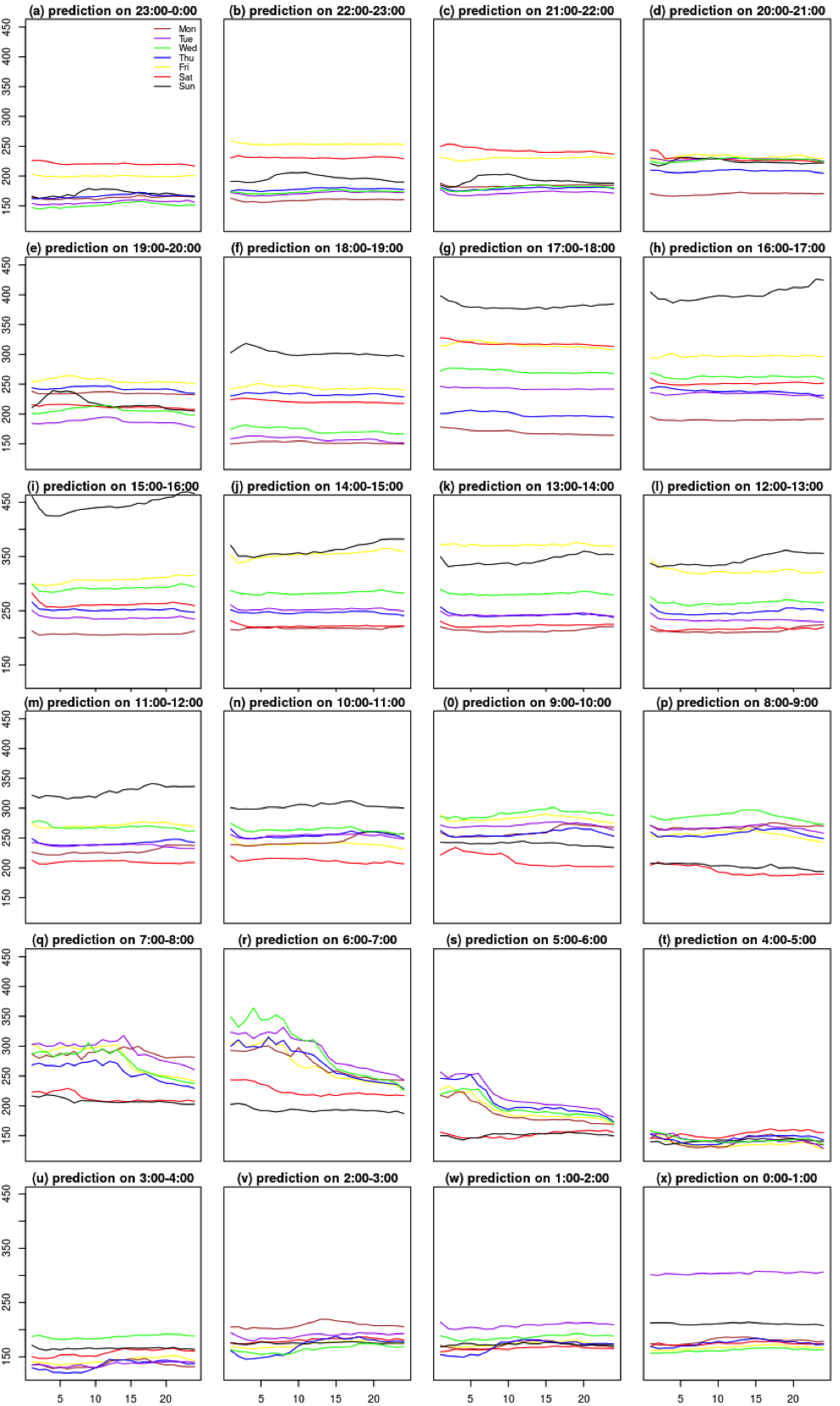}
\caption{Predictive accuracy ($y$-axis, measured in RMSE) of the 4,032 different XgBoost models corresponding to different 
days of the week (as indicated by different colors), hours of the day (header) and different amount of training data ($x$-axis, given in previous hours from $1$-$24$).} \label{fig4}
\end{figure}

Besides long-term predictions, short-term predictions for the next hours that react on recent conditions are needed. 
Here, an important question, especially with respect to storage and computation cost, is: how much information from the past is needed for training to obtain reasonable short predictions? To evaluate this, we trained several different XgBoost models, that only use the information from the last $i$ hours. Moreover, we compare the obtained predictions with the one calculated by using the long-term forecasts. To this end, we randomly selected the week from 2016-06-23 to 2016-06-29 and separately considered the trips from every hour of this week as a single test data set. This resulted in 168 (7 days $\times$ 24 hours/day) different test data sets. For each test data, we trained an XgBoost model model on the trips in its past $i$ hours, $i=1,\dots,24$ as described in the last section excluding the temporal features \emph{weekday} and all weather data. 
Thus, we trained 4,032 ($168 \times 24$) XgBoost models in total. Their performance is shown in Figure \ref{fig4}, in which each point represents a single model. 

Here, several things stand out: First, for most hours of day and days of the week, the RMSE was almost constant over the different amounts of training data ($1-24$ hours). This means 
that the trip duration of most hours and days have a rather short memory and mostly depend only on the trips from the last hour. An exception is given by the rush hours (5:00-8:00 on workdays), where the RMSE on test data was the lowest when using the past 24 hours as training data. Beside that, all models exhibit the best performance on early morning hours (1:00 - 4:00) with RMSEs around 150 for all hour $\times$ day combinations. However, on Sundays between 10:00 to 18:00, the trip duration was relatively hard to predict resulting in RMSEs around 450 or even larger. 
The average training times for XgBoost (CPU) are very low: They range from below one second for the case, in which only trips of the last hour were considered, to around 9 seconds for the case in which data of the last 24 hours were considered. 


\subsection{Short vs. Long-Term Prediction}
\begin{figure}
\includegraphics[width=\textwidth]{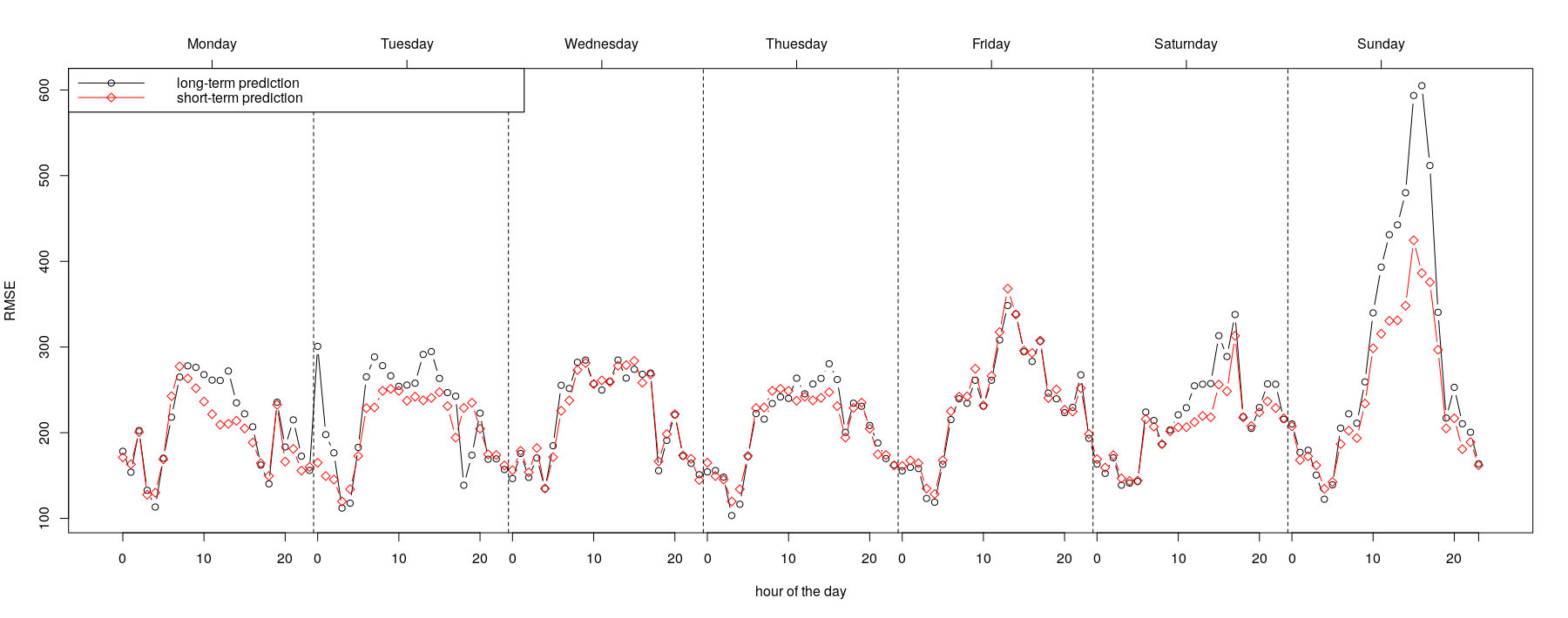}\vspace{-.5cm}
\caption{Comparison of the RMSE of the best CPU-based XgBoost long-term prediction model and the best short-term model.} \label{fig3}
\end{figure}

The long-term prediction models can in principle also be used to predict the travel time for the trips in this randomly selected week. Choosing, the best CPU-based XgBoost long-term prediction model (Table~\ref{tab_comptreebase}), we also computed its RMSE for each of 
the 168 different test data sets (hours of this specific week). A comparison with the best  short-term model is given in Figure~\ref{fig3}.
It is surprising, that there is not much difference for most of the time. In fact, the short-term models are slightly better than the long-term model from Monday to Thursday and even much better on Sundays. An explanation may be that short-term models react more directly to the current traffic condition which is more hidden in the much larger training set of the long-term model. Thus, if the aim is short-term prediction, the consequences are very positive and save a lot of computation time: Simply use the short-term models.

\section{Discussion and Outlook}

The results of our computational study show that ensemble tree methods deliver accurate travel time forecasts for both planning horizons, especially, if in addition to the core trip data, data related to the fastest route assuming a free flow speed, are considered. 
For short-term predictions, only data from the last few hours contain the information, which is necessary for a result outperforming models trained on a larger data set.
Due to the relatively small training data sets, the resulting training times are very short, and  allow for regular training runs on the most recent data.
That means, at the same time, that for different planning horizons different models should be provided.

In future work we want to investigate how much data is needed for reliable forecasts for both time horizons, since there are significantly more trip data available for New York than for most other cities in the world. 
Especially in cases with less data records, different ways of integrating the short-term state of the traffic might be promising such as the global average speed of the last last half an hour or the average speed of trips with similar origin destination locations. 
For many applications, it is also interesting to deliver arrival time (prediction) intervals or distributions expressing the uncertainty of the forecast.

\end{document}